# Trajectory Approximation of Video Based on Phase Correlation for Forward Facing Camera

Abdulkadhem A. Abdulkadhem
*College of Engineering - Department of Computer Technology Engineering -Al-Mustaqbal University, Babylon, Iraq.*

kazum2006k@yahoo.com

***Abstract*— In this paper, we introduce an innovative approach for extracting trajectories from a camera sensor in GPS-denied environments, leveraging visual odometry. The system takes video footage captured by a forward-facing camera mounted on a vehicle as input, with the output being a chain code representing the camera's trajectory. The proposed methodology involves several key steps. Firstly, we employ phase correlation between consecutive frames of the video to extract essential information. Subsequently, we introduce a novel chain code method termed "dynamic chain code," which is based on the x-shift values derived from the phase correlation. The third step involves determining directional changes (forward, left, right) by establishing thresholds and extracting the corresponding chain code. This extracted code is then stored in a buffer for further processing. Notably, our system outperforms traditional methods reliant on spatial features, exhibiting greater speed and robustness in noisy environments. Importantly, our approach operates without external camera calibration information. Moreover, by incorporating visual odometry, our system enhances its accuracy in estimating camera motion, providing a more comprehensive understanding of trajectory dynamics. Finally, the system culminates in the visualization of the normalized camera motion trajectory.**

## I. INTRODUCTION

Object tracking and trajectory estimation of vehicles using cameras are widely studied in the computer and robotic vision domain, with limited application in GPS-denied environments [1]. GPS may be absent or become less effective when only a few satellites are available in urban or remote areas. Though algorithms like Visual odometry (VO) and Simultaneous Localization and Mapping (SLAM) are used in outdoor localization and they require. Current expansions in image processing techniques and video capturing technologies support visual odometry localization with less computational overhead [2]. The important problems facing using track the trajectory of movement and shape of objects is hard to determine the appropriate method to describe this track variables[2]. The trajectory of an object observed in a video camera can be represented as a temporally sorted sequence of 2D-coordinates. The time interval between each point is constant and is determined by the frame rate of the camera [3].

Most trajectory extraction methods depend on the corresponding the extracting features from the spatial domain of frames of video. In these methods, extract important features such as corner points from the first frame and tracking points to the second frame for extracting the rotation matrix and translation vector between frames and then estimate the trajectory movement. The drawback of these methods are slow and can not be done without other parameters such as (intrinsic, camera, extrinsic camera, focal length, the principal point, distortion coefficients, ground truth locations, camera calibration). Thus, our proposed system focus on features extracted from the frequency domain of frames by using a phase correlation method and used these features to find chain code representation of the camera motion (trajectory).

The scenario of the proposed system is using monocular camera (single) setups as Forward-Facing Camera (FFC) of a vehicle or car. In figure(1) we notice, there are three possible movement to (forward, left, right) . The goal of the proposed system is to extract the trajectory of the vehicle and represented as a chain code based on video film only without any other information.

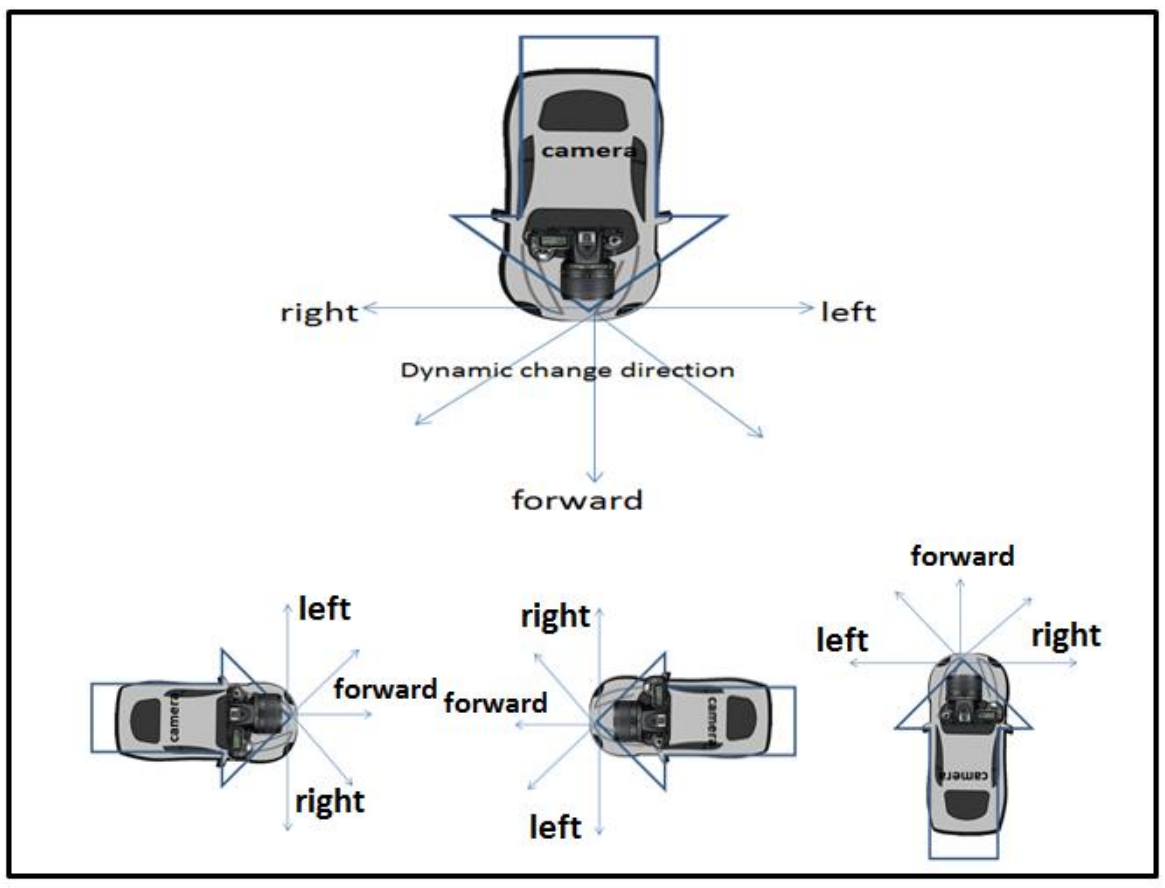

*Figure(1): The forward-facing camera.*

In the organization of paper, section II, Background, gives a thorough explanation of the chain code and phase correlation method. It is followed by the proposed system, which describes how the algorithms presented in section III. Section IV, Experimental Results, gives an explanation about the output trajectory of camera motion. In section V, Conclusions, where the authors reflect on the used method, achieved results and further improvements.

## II. BACKGROUND

This section explains the main idea of phase correlation approach and the traditional chain code method.

### A. Phase correlation

The phase correlation is a technique of cross-correlation based on the Fourier transform; because of the wide optimizations that fast Fourier transform (FFT) algorithms allow, phase correlation can be calculated very quickly. In its most straight forward implementation, phase correlation can be used to register together two images that differ by a relative translation and can be extended to differences in rotation and scaling. Phase correlation has been successfully applied to satellite imaging, augmented reality and motion estimation, featuring distinctive robustness against noise.[4]. The idea behind this registration method is based on the Fourier shift property, which states that a shift in the coordinate frames of two functions is transformed in the Fourier domain as a linear phase difference.[5]. The translation movement between two consecutive frames of video can be extracted using the phase correlation algorithm that represents the location of the maximum peak in the phase correlation map. figure(2) explains the phase correlation map between two consecutive frames.

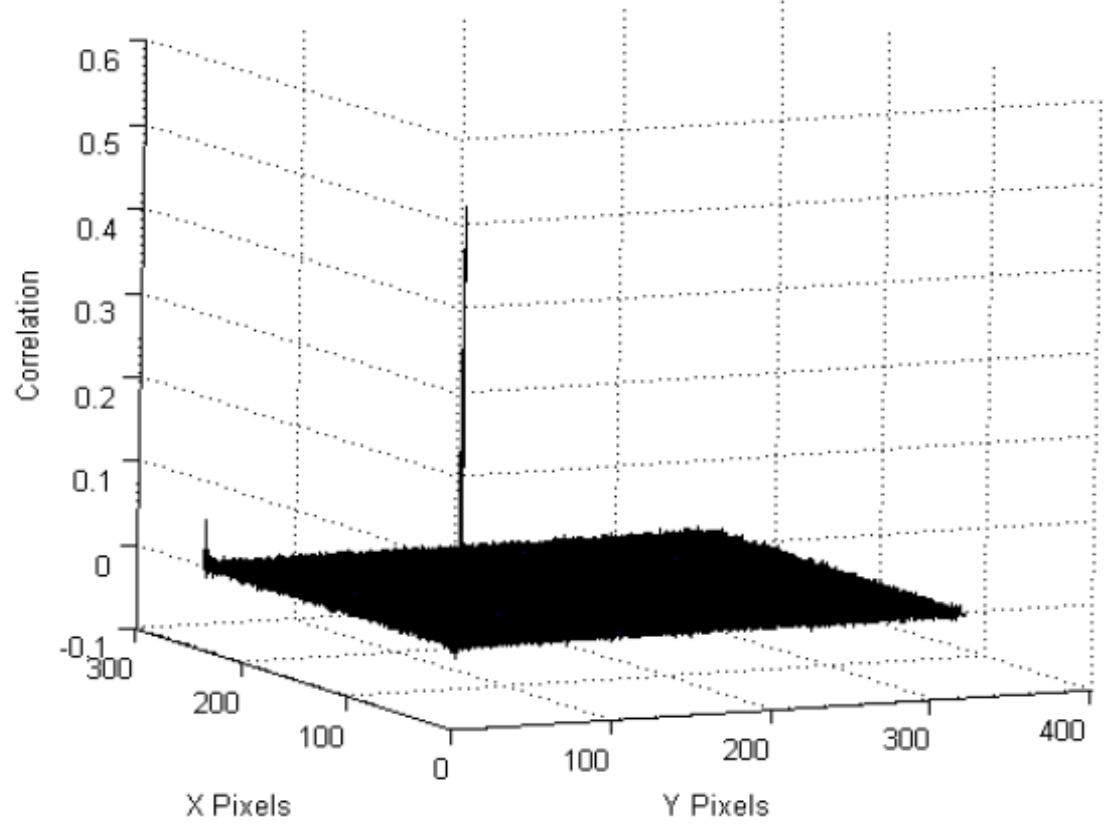

*Figure(2): The phase correlation map of two consecutive frames of video.*

### B. Chain code representations.

Freeman Chain Code (FCC) was the first technique to represent an image that uses chain code; it was introduced by Freeman in 1961. Straight-line segments that are connected in sequence with particular length and direction are represented as a boundary by using this chain code. This representation is based on 4- or 8-connectivity of the segments. A numbering scheme is used to code the direction of each segment. 4-connected Freeman Chain Code is shown in figure (3-a) while figure (3-b) shows 8-connected Freeman Chain Code of 8-directional (FCCE).

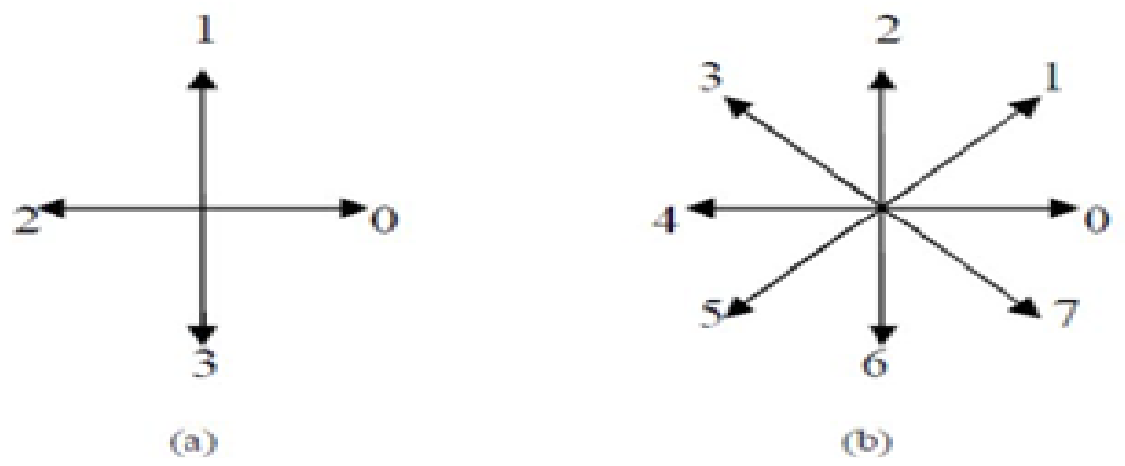

*Figure(3). The Neighbor directions of the freeman chain code.*

In the Freeman Chain Code 8-connected (FCCE), 8 directions from one pixel to a neighbor pixel are possible, every code is considered as an angular direction, multiplied by $45^0$ moving from one contour pixel to the next. Direction 0 mean move "to the right of", 2 means "immediately above", and 1 is at 45 degrees, bisecting 0 and 2, and so on.[6] in our proposed system will be using the adaptive 4-connectivity chain code for representing the trajectory of a video.

## III. THE PROPOSED SYSTEM

The proposed system as illustrated in Figure (4) consists of many steps.

The video is indeed the collection of images, each image is called the frame, displayed in a fast-paced enough so that the human eye can be aware of the continuity of its content. the image processing techniques can be applied to each frame. The contents of the sequential two frames are generally closely correlated [7]. The first step of the proposed system is the input video film file of a camera in an environment taking pictures(street) and take two consecutive frames. The others steps are explained as follow:

### A. The phase correlation method

Phase correlation method consists of the following operations:

1- Given 2 frames $\boldsymbol{g_a}$ *and* $\boldsymbol{g_b}$.

2- Calculate the discrete 2D Fourier Transform of both images: $\boldsymbol{G_a}=\boldsymbol{F}\{\boldsymbol{g_a}\}$,$\boldsymbol{G_b}=\boldsymbol{F}\{\boldsymbol{g_b}\}$.

3- Calculate the cross-power spectrum by multiplying the first Fourier transform and the complex conjugate of the second and normalizing the product element-wise.

$$\boldsymbol{R} = \frac{\boldsymbol{G_a} \circ \boldsymbol{G_b^*}}{|\boldsymbol{G_a} \circ \boldsymbol{G_b^*}|} \qquad (1)$$

4- Apply the inverse Fourier transform to obtain the normalized cross-correlation.

$$\boldsymbol{r}=\boldsymbol{F}^{-1}\{\boldsymbol{R}\} \qquad (2)$$

5- Determine the location of the peak in $\boldsymbol{r}$.[8]

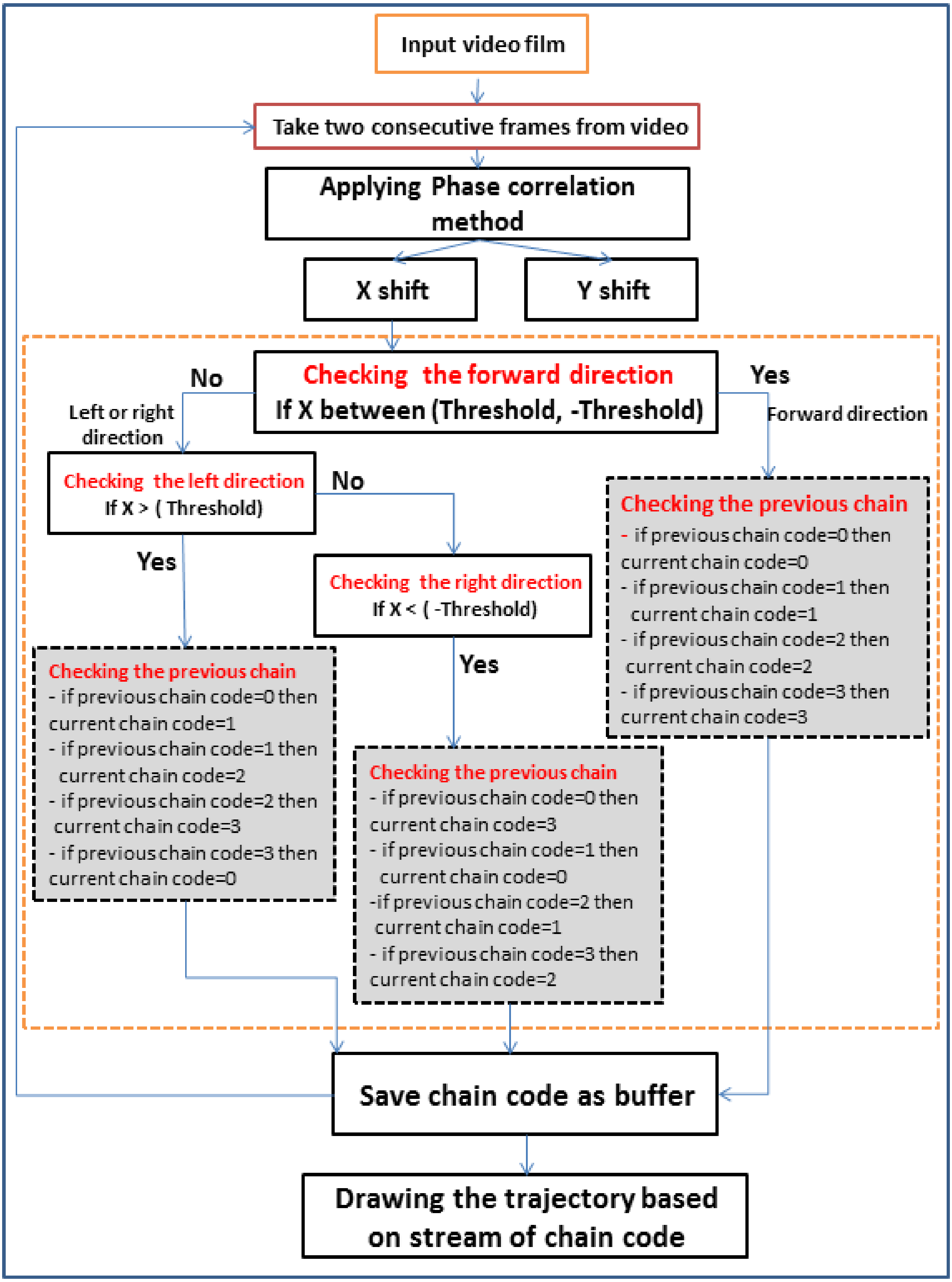

$$(\Delta x, \Delta y) = argmax_{(x,y)}\{r\} \quad (3)$$

*Figure(4). The block diagram of the proposed system*

the output of this step is the translation shift (x,y) between two consecutive frames of video. The important coordinate is the x-axis value that used to determine the direction change of camera moving. The value of the y-axis is not used in this paper.

*B. Chain code extraction*

The shift or movement between two consecutive frames of video can be represented as chain code. Each chain code have Initial State Chain Code (ISCC) and Movement State Chain Code (MSCC). In our proposed system , the 4-connectivity chain code is used ,thus, there are four possible true (actual) chain code called initial state (ISCC) that take values (0,1,2,3) as illustrated in figure (5-A). The value (0) represents the left direction. The value(1) represent the forward direction movement. The value (2) represents the right direction movement. And last, the value (3) represents a forward direction reverse the value (1). In figure(5-B) represent the dynamic chain code values that change over time according to the last direction extracted called the movement state chain code (MSCC). There are four possible values for the movement state chain code (0,1,2,3) that represent (left, forward, right, no change) respectively.

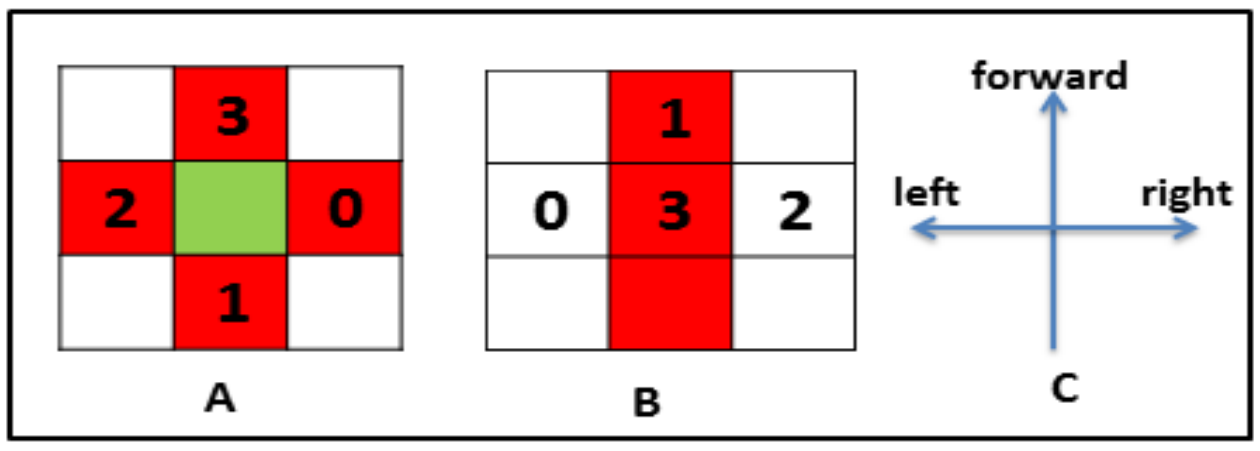

*Figure(5). A. The initial state chain code, B. The dynamic chain code (movement state), C. directions of camera movement*

In this step, there are three checking operations, the first one represent checking the forward moving direction of camera, the second one is checking the right moving direction, and lastly one is checking the left moving direction. The three checking operations are illustrated as follow:

1- The Forward direction.

The moving between two consecutive frames is forward when the value of x shift from above step between (-threshold, threshold). From experimental we note the value of the threshold is (12-60). In the beginning, the first chain code of the first two consecutive frames is forward moving (ISCC =1 , MSCC=1). The current chain code extracted depend on the previous initial state chain code (ISCC). In this operation the current chain code equal to the previous chain code. Figure (6) explains the forward direction checking movement.

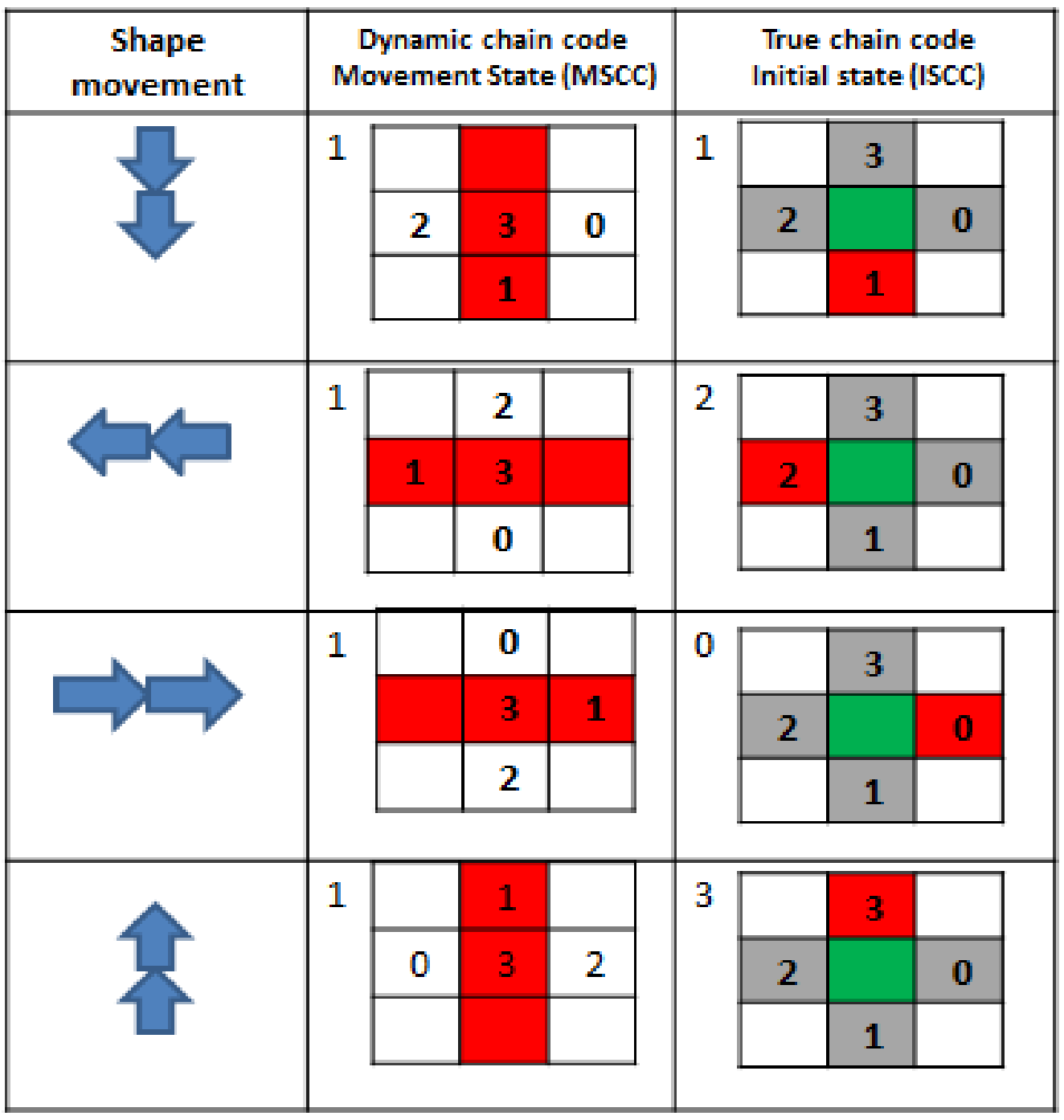

*Figure(6). The forward direction checking.*

2- The left direction

The current chain code extracted depend on the previous initial state chain code (ISCC) . In this operation the current chain code is equal to the rotated previous chain code by 90 degree in counterclockwise direction. For example, if the previous chain code representation is forward of right direction (true chain code (ISCC) =0, dynamic chain code (MSCC) =1) then the current dynamic chain code (MSCC) will be (0) and the (actual) true chain code (ISCC) will be (3). Figure (7) explains the left direction checking movement. In this step will be skipping the next compute the actual chain code until till the forward direction movement, that means the value of x shift of the phase correlation must be less than a threshold.

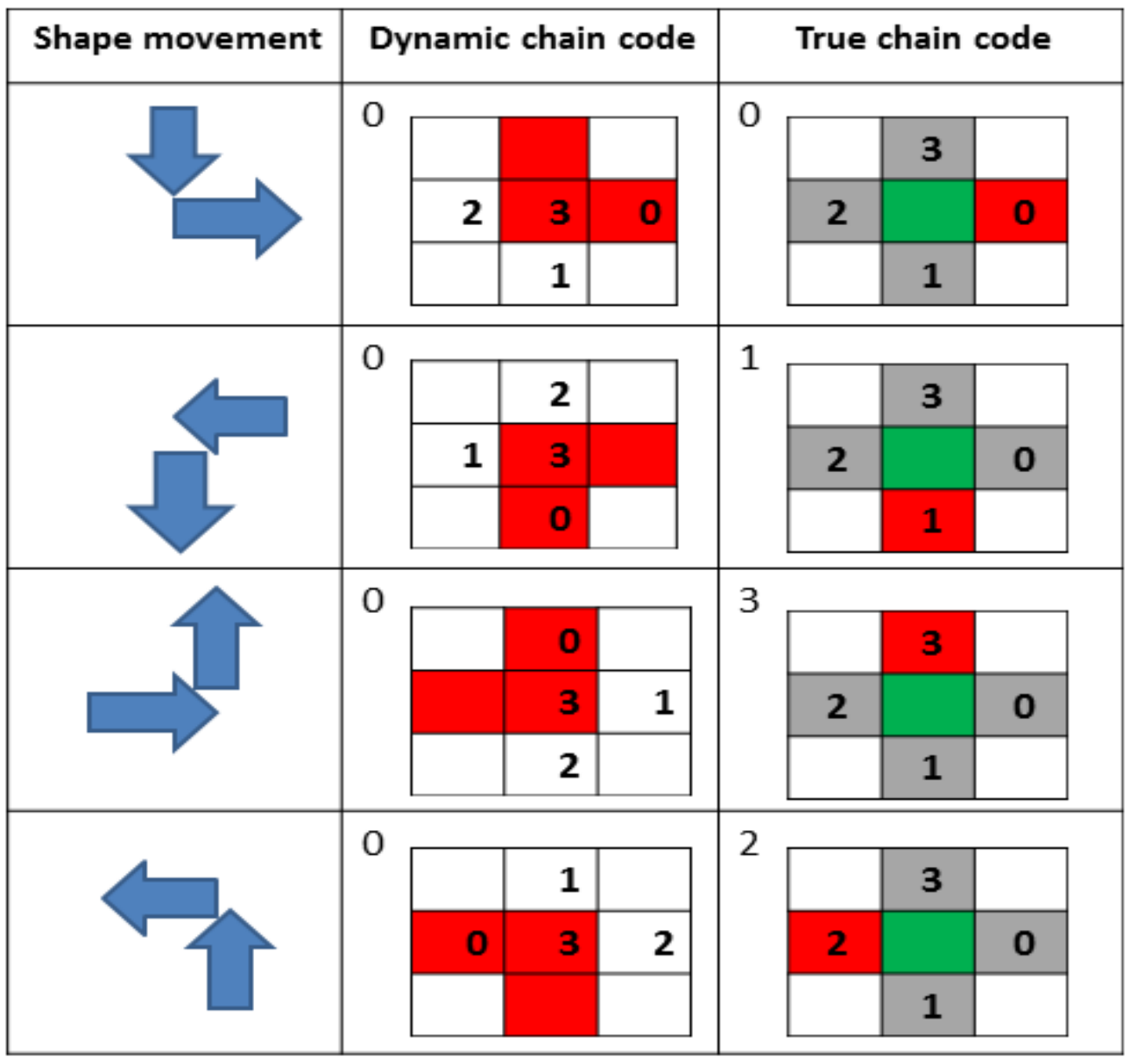

*Figure(7). The left direction checking.*

3- The right direction

The current chain code extracted depend on the previous actual chain code (ISCC). In this operation the current chain code is equal to the rotated previous chain code by 90 degree in clockwise direction. For example, if the previous chain code representation is forward of left direction(true chain code (ISCC) =2, dynamic chain code (MSCC) =1) then the current dynamic chain code (MSCC) will be (2) and the true chain code (ISCC) will be (3). Figure (8) explains the left direction checking movement. In this step will be skipping the next compute the actual chain code until till the forward direction movement, that means the value of x shift of the phase correlation must be larger than (-threshold).

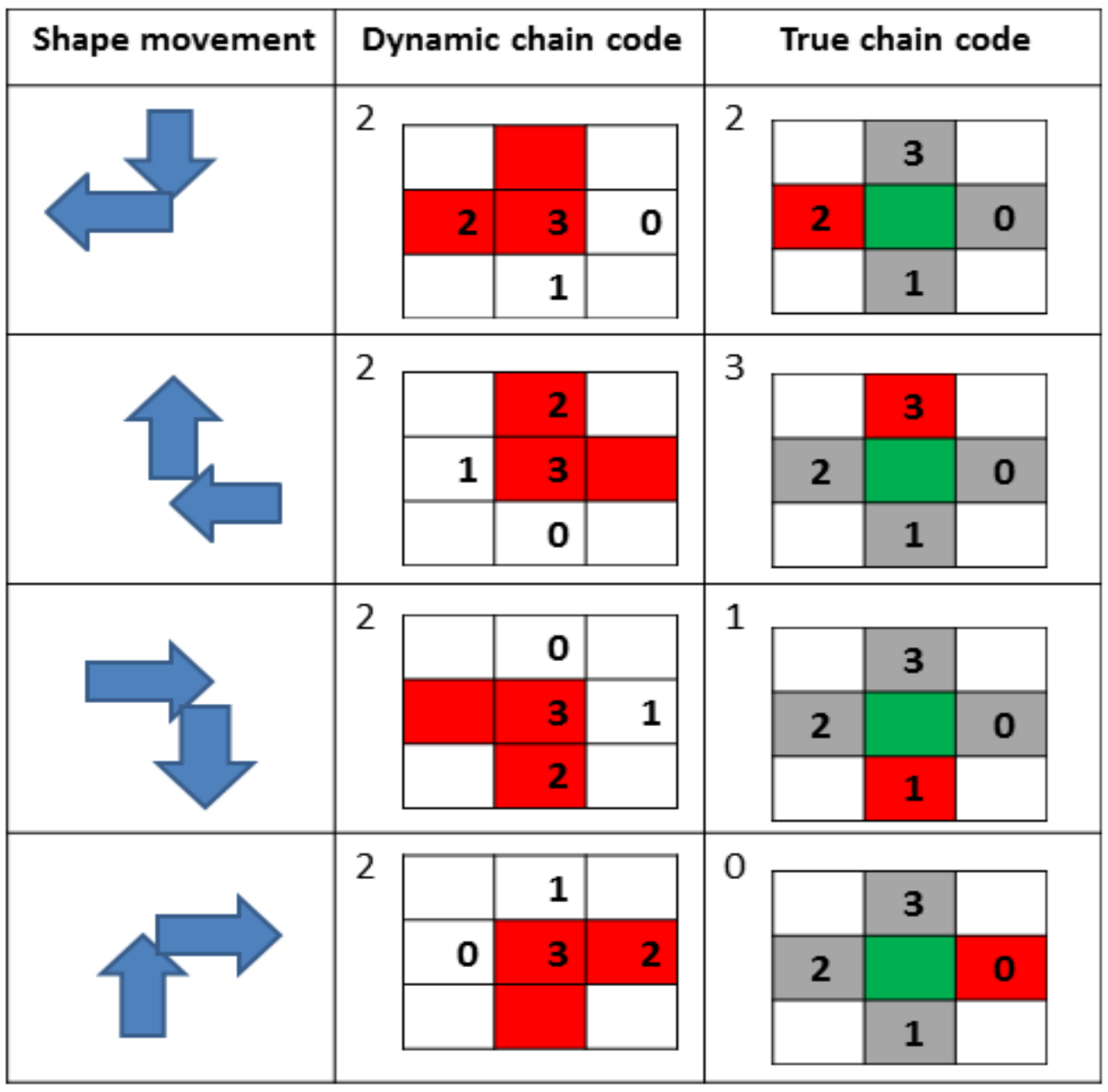

*Figure(8). The left direction checking.*

### *C. Saving the chain code in the buffer*

In this step will be saving the actual chain code (ISCC) that represents the moving direction between two consecutive frames of video film. The next step is taking the new consecutive frames from video and applying the above steps and operations until process the last consecutive frames of a video.

### *D. Drawing trajectory of video*

The last step of the proposed system is drawing the trajectory based on the chain code according to the shape of the figure(3-A). The output of this step is a trajectory simplified. The stream of the chain code used to determine the direction of trajectory. The constant distance used between each chain code when drawing the trajectory.

It’s important to notice the length of the actual chain code (ISCC) is smaller than the number of frames of the video because of many frames are skipped in an abrupt change in direction to the left or right direction when computing the movement states chain code (MSCC). The length of the dynamic chain code (MSCC) is equal to the frames numbers of video minus one.

## IV. EXPERIMENTAL RESULTS

Experiments of the proposed method carried out to prove the efficiency. The proposed method has been simulated using vb.net program on the Windows 7 platform on Intel Core i5 2.5 GHz with 4 GB of main memory. Experiments of the proposed method in this paper are performed on two sets of sequential frames of video film taken from the Kitti vision benchmark[9]. The frame dimensions are (1267x387) pixels. The type of camera used in the proposed system is a monocular camera (the single camera). The first set sequence frames are (241) frames as illustrated below in figure(9)

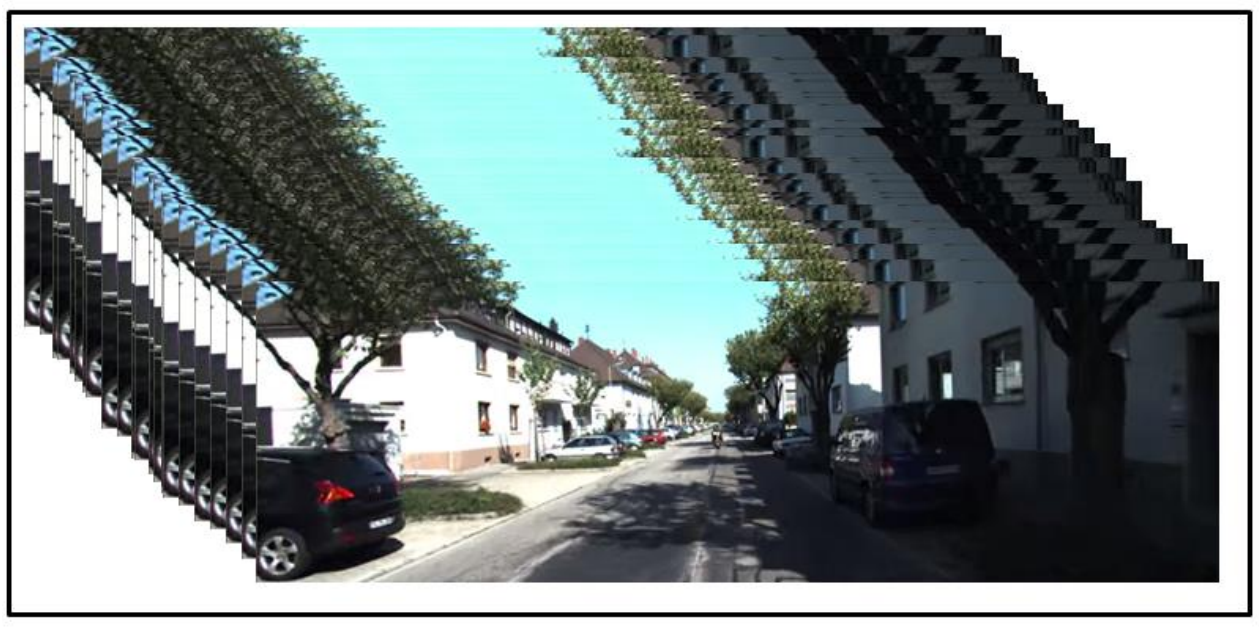

*figure(9). The first set of frames video.*

After applying the proposed system to the first study case sequence frames, when the threshold is (16), the length of dynamic chain code or movement state chain code (MSCC) is (240) chain code, while the length of actual chain code or initial state chain code(ISCC) is (215) chain code because there are some frames are skipped from the movement state in the values of x-axis of phase correlation not check the condition (x shift must between –threshold and threshold). In the table(I) illustrated the stream of dynamic movement state chain code and actual initial state chain code that characterize the trajectory representation of video[9,10,11].

TABLE I. THE DYNAMIC AND ACTUAL CHAIN CODE REPRESENTATION OF A TRAJECTORY FOR THE FIRST SET OF FRAMES

| Dynamic chain code (MSCC) | Actual chain code (ISCC) |
|---|---|
| 1,1,1,1,1,1,1,1,1,1,1,1,1,1,1,1,1,1,1<br>,1,1,1,1,2,3,3,3,3,3,3,3,1,1,1,1,1,1<br>,1,1,1,1,1,1,1,1,1,1,1,1,1,1,1,1,1,1<br>,1,1,1,1,1,1,1,1,1,1,1,1,1,1,1,1,1,<br>1,1,1,1,1,1,1,1,1,1,1,1,1,1,1,1,1,1<br>,1,1,1,1,1,1,1,1,1,2,3,3,3,3,3,3,3,3,<br>1,1,1,1,1,1,1,1,1,1,1,1,1,1,1,1,1,1,<br>1,1,1,1,1,1,1,1,1,1,1,1,1,1,1,1,1,1<br>,1,1,1,1,1,1,1,1,1,1,1,1,1,1,1,1,1,1<br>,1,1,1,1,1,1,1,1,1,1,1,1,1,1,1,1,1,<br>1,1,1,1,1,1,1,1,1,1,1,1,1,1,1,1,1,1<br>,1,1,1,1,1,1,1,1,1,1,1,1,1,1,1,1,1,<br>1,1,1,1,1,1,1,1,1,1,1,1,1,1,1,1,1,1<br>,1,1,1,1,1, | 1,1,1,1,1,1,1,1,1,1,1,1,1,1,1,1,1,1,1,<br>1,1,1,1,2,2,2,2,2,2,2,2,2,2,2,2,2,2,2<br>,2,2,2,2,2,2,2,2,2,2,2,2,2,2,2,2,2,2,2<br>,2,2,2,2,2,2,2,2,2,2,2,2,2,2,2,2,2,2,2<br>2,2,2,2,2,2,2,2,2,2,2,3,3,3,3,3,3,3,3,<br>3,3,3,3,3,3,3,3,3,3,3,3,3,3,3,3,3,3,3,<br>3,3,3,3,3,3,3,3,3,3,3,3,3,3,3,3,3,3,3<br>,3,3,3,3,3,3,3,3,3,3,3,3,3,3,3,3,3,3,<br>3,3,3,3,3,3,3,3,3,3,3,3,3,3,3,3,3,3,3<br>,3,3,3,3,3,3,3,3,3,3,3,3,3,3,3,3,3,3,<br>3,3,3,3,3,3,3,3,3,3,3,3,3,3,3,3,3,3,3,<br>3,3,3,3,3,3,3,3,3,3 |

Figure(10) illustrates the movement state chain code of the first set frames video that represent the dynamic chain code method when the window changes the direction with the same direction of camera motion .

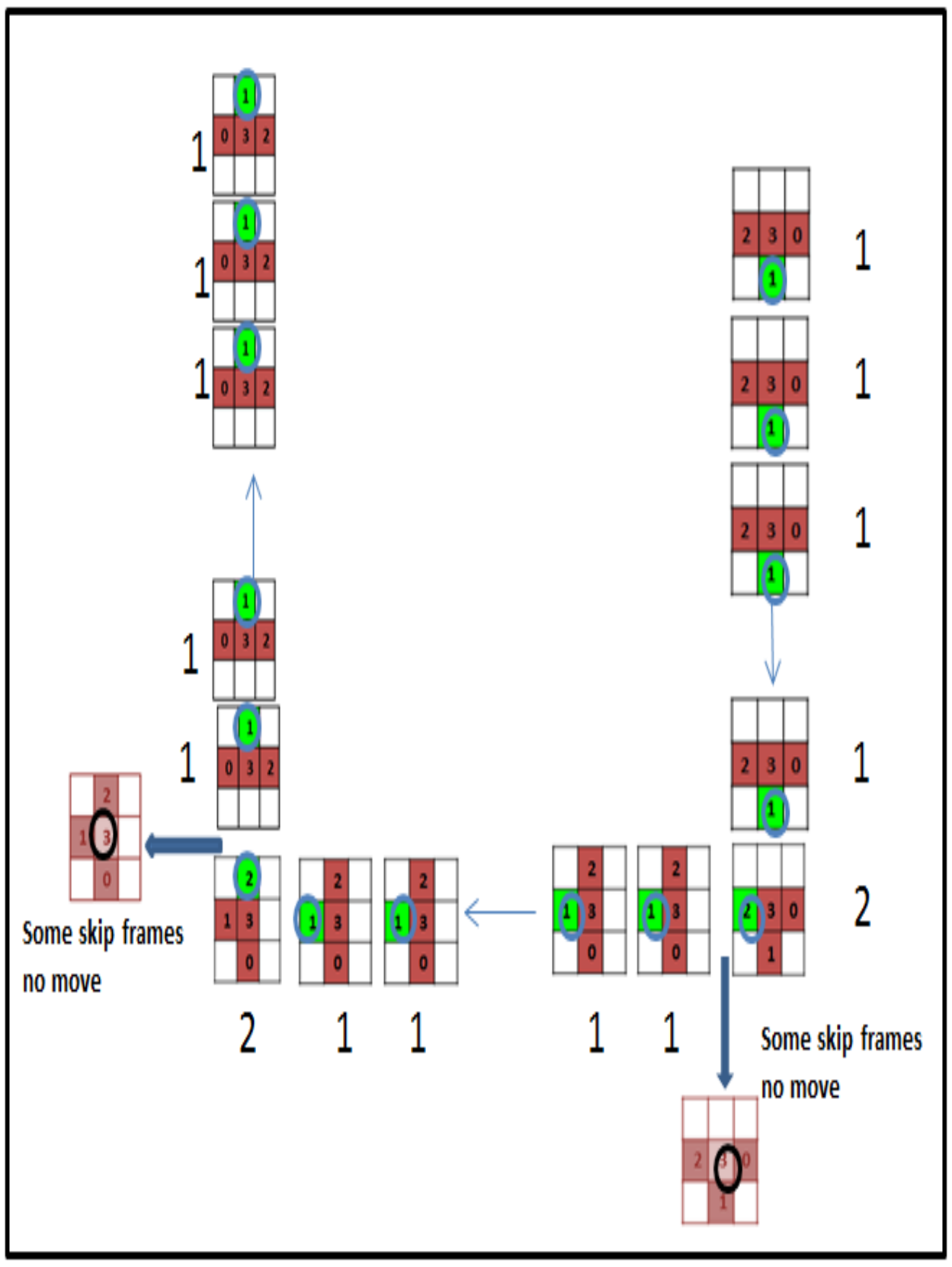

*figure(10). The movement state chain code of the first set frames video.*

Figure(11) illustrates the initial state chain code of the first set frames video that represent the actual chain code representation of trajectory for camera motion .

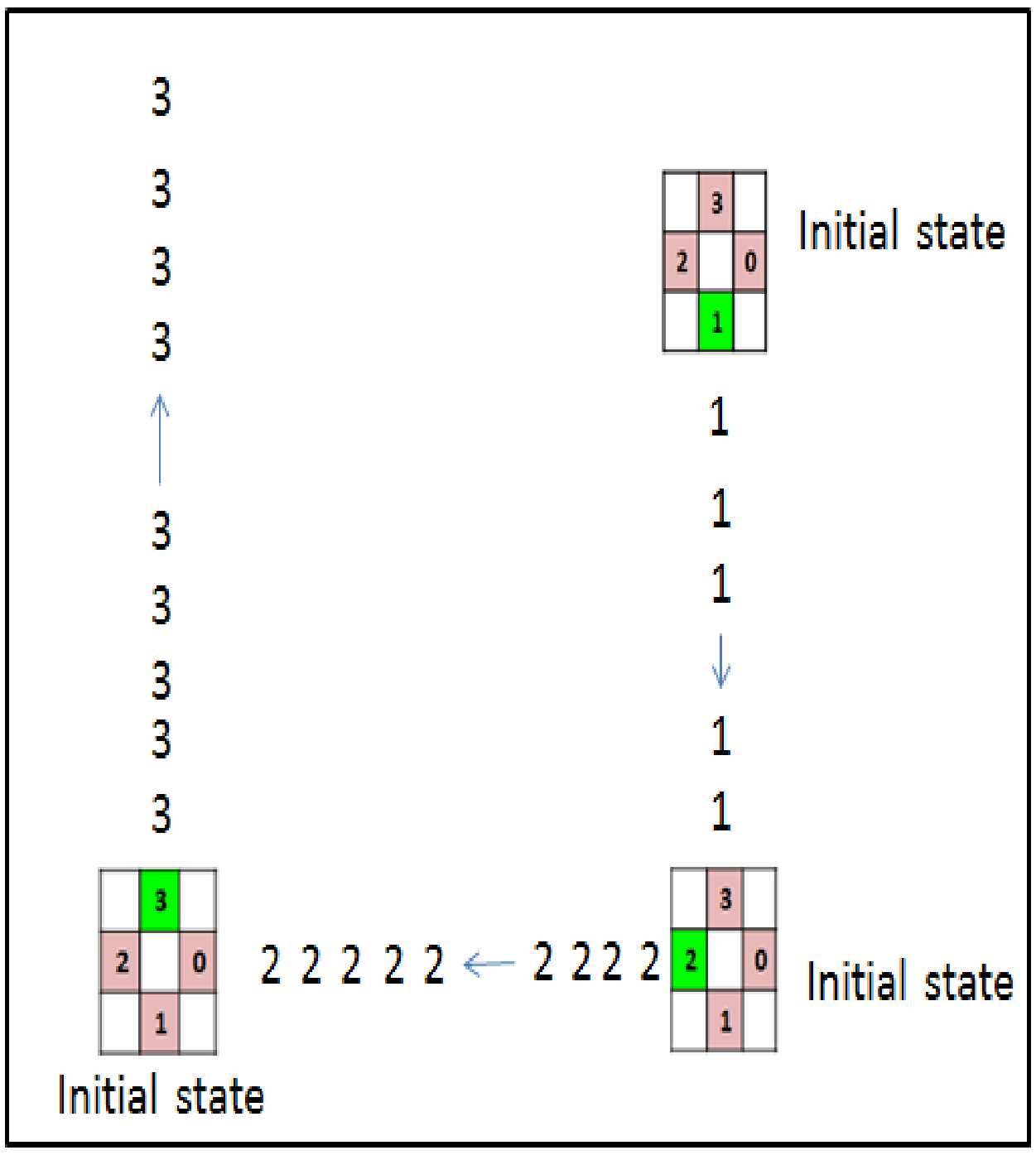

*figure(11). The Initial state chain code of the first set frames video.*

Figure(12) illustrates the ground truth trajectory and the proposed extracted trajectory representation of the camera motion of the first set frames video.

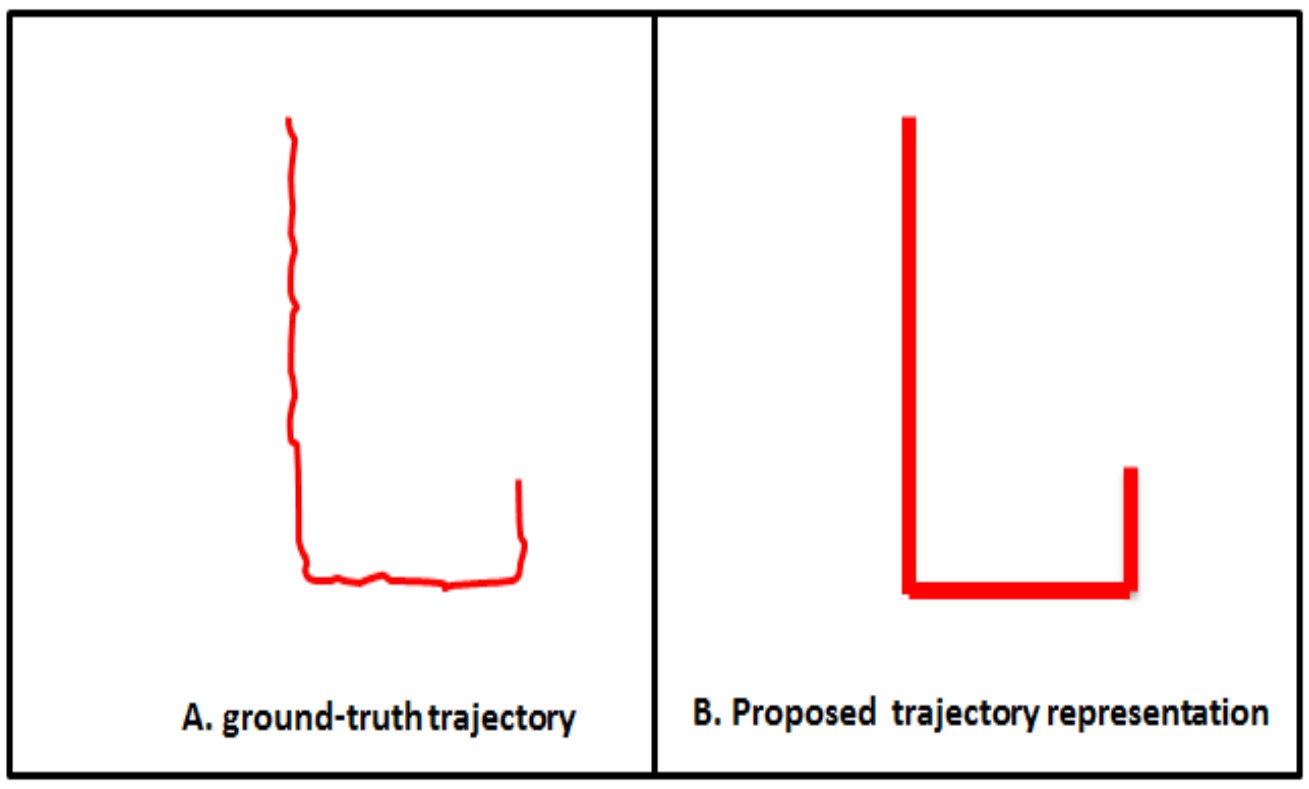

*Figure(12). Trajectory representation of the first set frames video. A. the ground truth motion trajectory, B. the proposed trajectory representation.*

The second case study is (200) sequence frames as illustrated in figure(13).

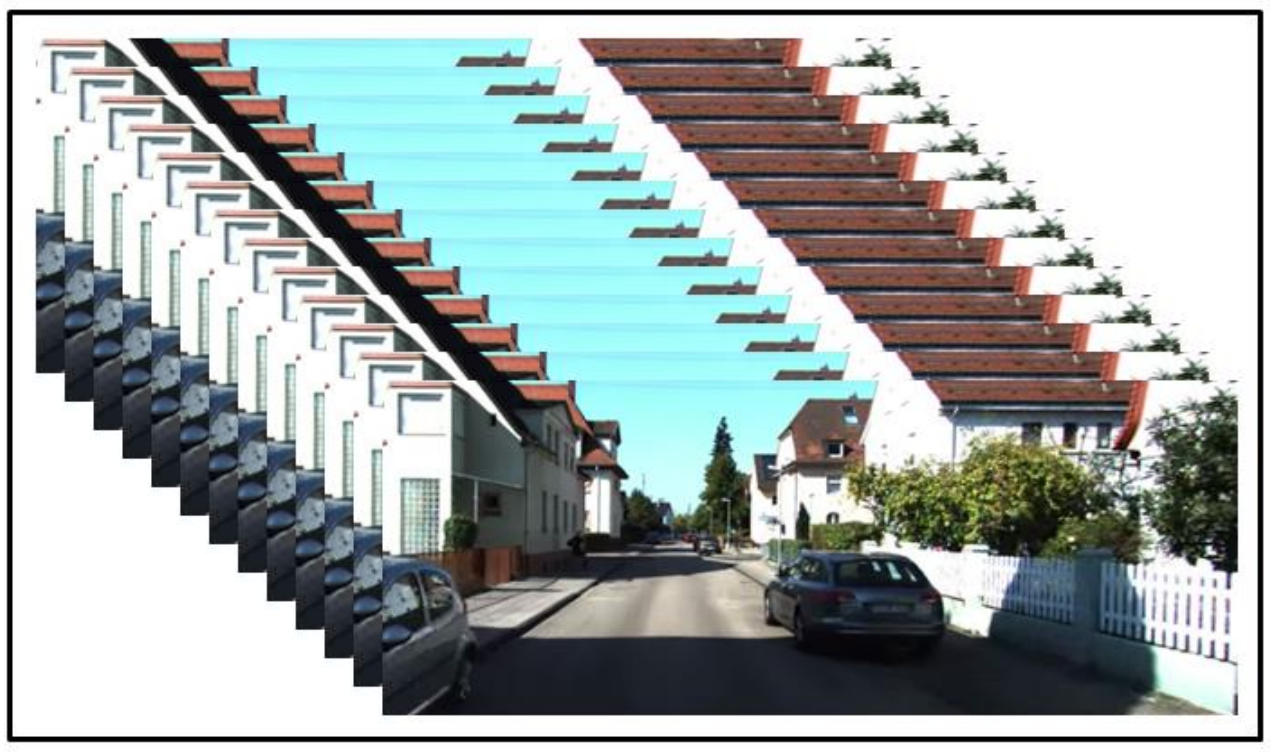

*figure(13): The second set of frames.*

After applying the proposed system to the second study case sequence frames, when the threshold is (20), the length of the dynamic chain code (MSCC) is (199) chain code, while the length of the actual chain code (ISCC) is (176) chain code because there are some frames are skipped from the actual chain code in the values of x-axis of phase correlation not check the condition (x shift must between –threshold and threshold). In the table(II) illustrated the stream of dynamic chain code and actual chain code that characterize the trajectory representation of video.

TABLE II. THE DYNAMIC AND ACTUAL CHAIN CODE REPRESENTATION OF A TRAJECTORY FOR THE SECOND SET OF FRAMES

| Dynamic chain code | Actual chain code |
|---|---|
| 1,1,1,1,1,1,1,1,1,1,1,1,1,1,1,1,1,1,1<br>,1,1,1,1,0,3,3,3,3,3,3,3,1,1,1,1,1,1<br>,1,1,1,1,1,1,1,1,1,1,1,1,1,1,1,1,1,1<br>,1,1,1,1,1,1,1,1,1,1,1,1,1,1,1,1,1,1,<br>1,1,1,1,1,1,1,1,1,1,1,1,1,1,1,1,1,1<br>,1,1,1,1,1,1,1,1,1,2,3,3,3,3,3,3,3,3,<br>1,1,1,1,1,1,1,1,1,1,1,1,1,1,1,1,1,1,<br>1,1,1,1,1,1,2,3,3,3,3,3,3,3,3,1,1,1<br>,1,1,1,1,1,1,1,1,1,1,1,1,1,1,1,1,1,1<br>,1,1,1,1,1,1,1,1,1,1,1,1,1,1,1,1,1,1,<br>1,1,1,1,1,1,1,1,1,1,1,1,1,1,1,1,1,1,<br>1 | 1,1,1,1,1,1,1,1,1,1,1,1,1,1,1,1,1,1,<br>1,1,1,1,1,0,0,0,0,0,0,0,0,0,0,0,0,0<br>,0,0,0,0,0,0,0,0,0,0,0,0,0,0,0,0,0,0<br>,0,0,0,0,0,0,0,0,0,0,0,0,0,0,0,0,0,0<br>,0,0,0,0,0,0,0,0,0,0,0,0,0,1,1,1,1<br>,1,1,1,1,1,1,1,1,1,1,1,1,1,1,1,1,1,1<br>,1,1,2,2,2,2,2,2,2,2,2,2,2,2,2,2,2,2<br>,2,2,2,2,2,2,2,2,2,2,2,2,2,2,2,2,2,2<br>,2,2,2,2,2,2,2,2,2,2,2,2,2,2 |

Figure(14) illustrates the movement state chain code of the second set frames video that represent the dynamic chain code method when the window changes the direction with the same direction of camera motion .

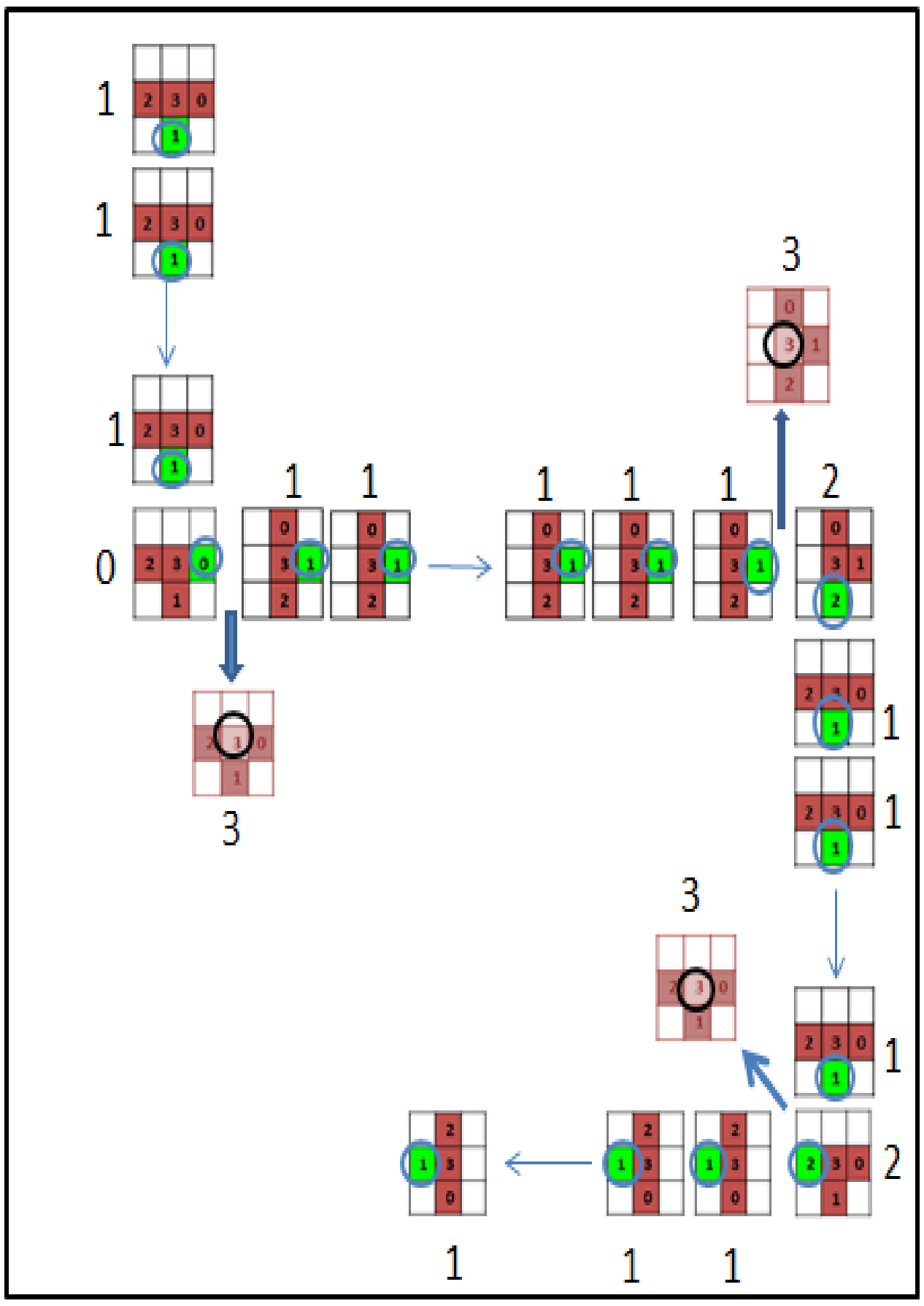

*figure(14). The movement state chain code of the second set frames video.*

Figure(15) illustrates the initial state chain code of the second set frames video that represent the actual chain code representation of trajectory for camera motion .

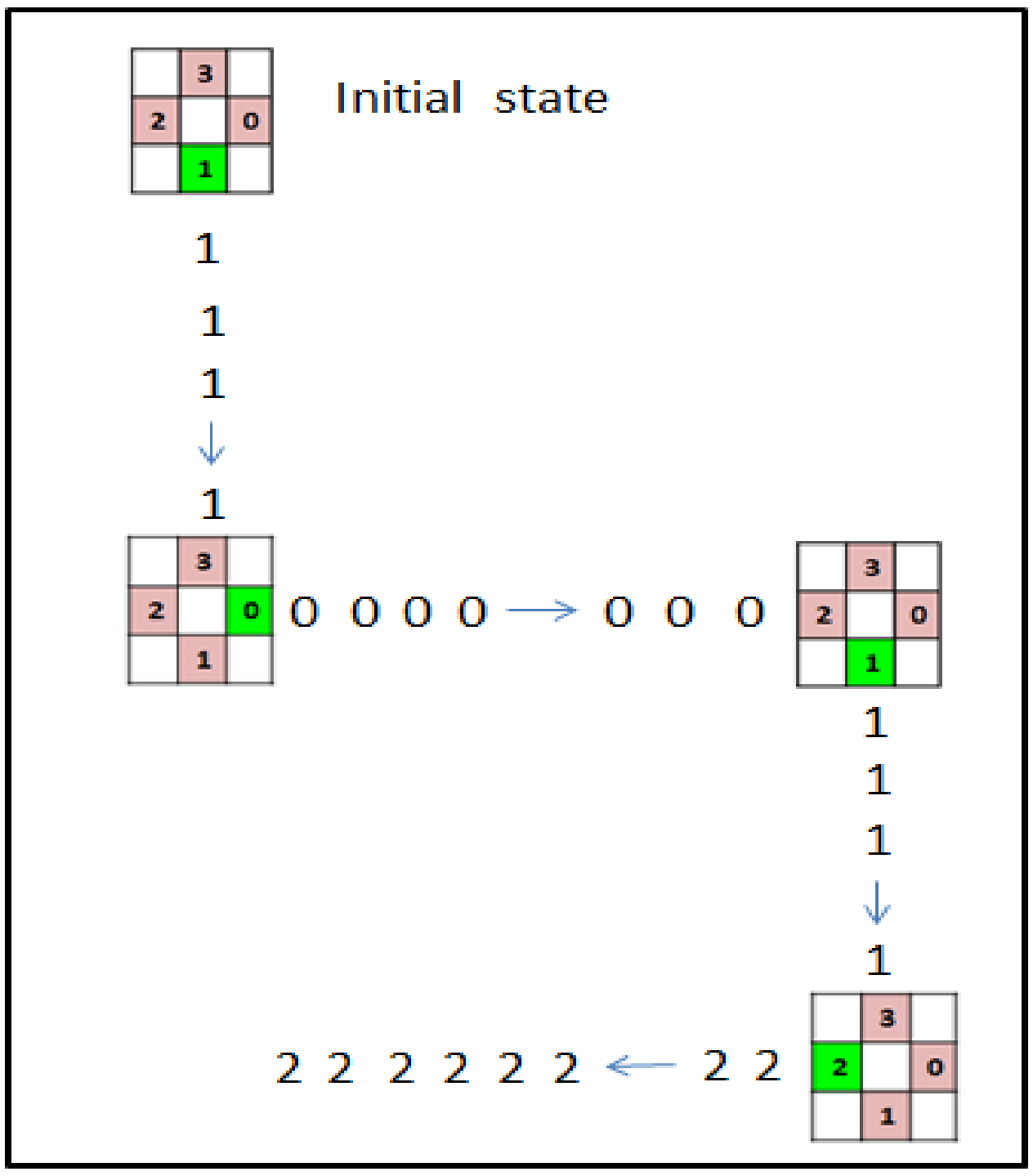

*figure(15). The Initial state chain code of the second set frames video.*

Figure(16) illustrates the ground truth trajectory and the proposed extracted trajectory representation of the camera motion of the second set frames video.

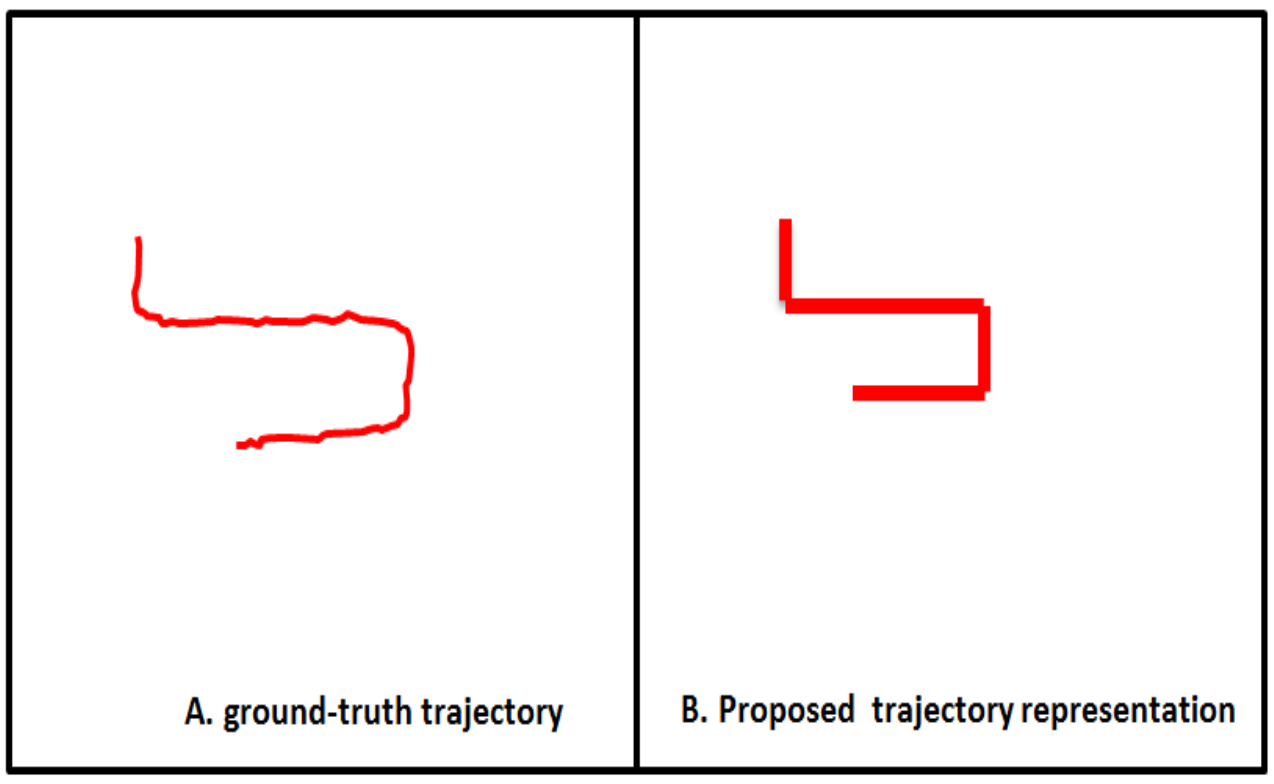

*Figure(16). Trajectory representation of the second set frames video. A. the ground truth motion trajectory, B. the proposed trajectory representation.*

## V. CONCLUSIONS

In summary, our study successfully implemented an efficient system for trajectory estimation using video footage from a forward-facing vehicle-mounted camera. The introduction of a dynamic chain code for trajectory extraction proved effective in normalizing and simplifying trajectories compared to ground truth GPS measurements. The system demonstrated a novel approach to trajectory representation, utilizing a compact two-bit code for each frame's movement. Unlike traditional methods relying on local features, our dynamic chain code leveraged global features, enhancing robustness. Future research may explore real-time implementation, extended environmental testing, sensor fusion integration, machine learning, and scalability for broader applications.